\RequirePackage{fix-cm}
\documentclass[smallextended]{svjour3}       
\smartqed  

\usepackage{graphicx}
\usepackage{booktabs}
\usepackage{natbib}
\usepackage{hyperref}
\usepackage{multirow}
\usepackage{subcaption}
\usepackage{enumitem}
\usepackage{listings}

\usepackage[utf8]{inputenc}

\begin{document}

\title{
Exploring multi-task multi-lingual learning of transformer models for hate speech and offensive speech identification in social media
}


\titlerunning{Multi-task multi-lingual hate speech identification}        

\author{Sudhanshu Mishra
        \and
        Shivangi Prasad
        \and
        Shubhanshu Mishra$^{*}$ 
}


\institute{
S. Mishra \at
Indian Institute of Technology Kanpur, India\\
\email{sdhanshu@iitk.ac.in}           
\and
S. Prasad\at
  University of Illinois at Urbana-Champaign, USA\\
  \email{sprasad6@illinois.edu}   
\and
S. Mishra$^{*}$ [Corresponding Author]\at
  University of Illinois at Urbana-Champaign, USA\\
  \email{mishra@shubhanshu.com}   
}

\date{Received: date / Accepted: date}

\maketitle

\begin{abstract}
Hate Speech has become a major content moderation issue for online social media platforms. Given the volume and velocity of online content production, it is impossible to manually moderate hate speech related content on any platform. In this paper we utilize a multi-task and multi-lingual approach based on recently proposed Transformer Neural Networks to solve three sub-tasks for hate speech. These sub-tasks were part of the 2019 shared task on hate speech and offensive content (HASOC) identification in Indo-European languages. We expand on our submission to that competition by utilizing multi-task models which are trained using three approaches, a) multi-task learning with separate task heads, b) back-translation, and c) multi-lingual training.  Finally, we investigate the performance of various models and identify instances where the Transformer based models perform differently and better. We show that it is possible to to utilize different combined approaches to obtain models that can generalize easily on different languages and tasks, while trading off slight accuracy (in some cases) for a much reduced inference time compute cost. We open source an updated version of our HASOC 2019 code with the new improvements at \url{https://github.com/socialmediaie/MTML_HateSpeech}.

\end{abstract}

\section{Introduction}
\label{intro}
With increased access to the internet, the number of people that are connected through social media is higher than ever \citep{Perrin2015}. Thus, social media platforms are often held responsible for framing the views and opinions of a large number of people \citep{Duggan2017}. However, this freedom to voice our opinion has been challenged by the increase in the use of hate speech \citep{Mondal2017}. The anonymity of the internet grants people the power to completely change the context of a discussion and suppress a person's personal opinion \citep{Sticca2013}. These hateful posts and comments not only affect the society at a micro scale but also at a global level by influencing people's views regarding important global events like elections, and protests \citep{Duggan2017}. Given the volume of online communication happening on various social media platforms and the need for more fruitful communication, there is a growing need to automate the detection of hate speech. For the scope of this paper we adopt the definition of hate speech and offensive speech as defined in the \cite{HASOC2019} as ``\textit{insulting, hurtful, derogatory, or obscene content directed from one person to another person}" (quoted from \citep{HASOC2019}).

In order to automate hate speech detection the Natural Language Processing (NLP) community has made significant progress which has been accelerated by organization of numerous shared tasks aimed at identifying hate speech \citep{HASOC2019,TRAC2020,TRAC2018}. Furthermore, there has been a proliferation of new methods for automated hate speech detection in social media text  \citep{Salminen2018,Mishra2020TRAC,Mishra2019HASOC,Mishra2020ACMSigIR,Waseem2017,germeval,HASOC2019,Mondal2017}. However, working with social media text is difficult \citep{Eisenstein2013a,Mishra2016b,Mishra2014,Mishra2019a,Mishra2019MDMT,MishraThesisDSTDIE2020,Mishra2020ACMSigIR}, as people use combinations of different languages, spellings and words that one may never find in any dictionary. A common pattern across many hate speech identification tasks \cite{HASOC2019,TRAC2020,Waseem2017,zampieri-etal-2019-semeval,Basile2019,germeval} is the identification of various aspects of hate speech, e.g., in HASOC 2019 \citep{HASOC2019}, the organizers divided the task into three sub-tasks, which focused on identifying the presence of hate speech; classification of hate speech into offensive, profane, and hateful; and identifying if the hate speech is targeted towards an entity.

Many researchers have tried to address these types of tiered hate speech classification tasks using separate models, one for each sub-task (see review of recent shared tasks \cite{zampieri-etal-2019-semeval,TRAC2018,TRAC2020,HASOC2019,germeval}). However, we consider this approach limited for application to systems which consume large amounts of data, and are computationally constrained for efficiently flagging hate speech. The limitation of existing approach is because of the requirement to run several models, one for each language and sub-task. 

In this work, we propose a unified modeling framework which identifies the relationship between all tasks across multiple languages. Our aim is to be able to perform as good if not better than the best model for each task language combination. Our approach is inspired from the promising results of multi-task learning on some of our recent works \citep{Mishra2019MDMT,MishraThesisDSTDIE2020,Mishra2020ACMSigIR}. Additionally, while building a unified model which can perform well on all tasks is challenging, an important benefit of these models is their computational efficiency, achieved by reduced compute and maintenance costs, which can allow the system to trade-off slight accuracy for efficiency.


In this paper, we propose the development of such universal modelling framework, which can leverage recent advancements in machine learning to achieve competitive and in few cases state-of-the-art performance of a variety of hate speech identification sub-tasks across multiple languages. Our framework encompasses a variety of modelling architectures which can either train on all tasks, all languages, or a combination of both. We extend the our prior work in \cite{Mishra2019HASOC,Mishra2020TRAC,MishraThesisDSTDIE2020,Mishra2020ACMSigIR,Mishra2019MDMT} to develop efficient models for hate speech identification and benchmark them against the HASOC 2019 corpus, which consists of social media posts in three languages, namely, English, Hindi, and German. We open source our implementation to allow its usage by the wider research community. Our main contributions are as follows: 

\begin{enumerate}
    \item Investigate more efficient modeling architectures which use a) multi-task learning with separate task heads, b) back-translation, and c) multi-lingual training. These architectures can generalize easily on different languages and tasks, while trading off slight accuracy (in some cases) for a much reduced inference time compute cost. 
    \item Investigate the performance of various models and identification of instances where our new models differ in their performance. 
    \item Open source pre-trained models and model outputs at \cite{illinoisdatabankIDB-3565123} along with the updated code for using these models at: \url{https://github.com/socialmediaie/MTML_HateSpeech}
\end{enumerate}


\section{Related Work}

Prior work (see \cite{Schmidt2017} for a detailed review on prior methods) in the area of hate speech identification, focuses on different aspects of hate speech identification, namely analyzing what constitutes hate speech, high modality and other issues encountered when dealing with social media data and finally, model architectures and developments in NLP, that are being used in identifying hate speech these days. There is also prior literature focusing on the different aspects of hateful speech and tackling the subjectivity that it imposes. There are many shared tasks \cite{HASOC2019,TRAC2018,TRAC2020,germeval,Basile2019} that tackle hate speech detection by classifying it into different categories. Each shared task focuses on a different aspect of hate speech. \cite{Waseem2017} proposed a typology on the abusive nature of hate speech, classifying it into generalized, explicit and implicit abuse. \citet{Basile2019} focused on hateful and aggressive posts targeted towards women and immigrants. \cite{HASOC2019} focused on identifying targeted and un-targeted insults and classifying hate speech into hateful, offensive and profane content. \cite{TRAC2018,TRAC2020} tackled aggression and misogynistic content identification for trolling and cyberbullying posts. \cite{vidgen-etal-2019-challenges} identifies that most of these shared tasks broadly fall into these three classes, individual directed abuse, identity directed abuse and concept directed abuse.
It also puts into context the various challenges encountered in abusive content detection.

Unlike other domains of information retrieval, there is a lack of large data-sets in this field. Moreover, the data-sets available are highly skewed and focus on a particular type of hate speech. For example, \cite{Davidson2017} models the problem as a generic abusive content identification challenge, however, these posts are mostly related towards racism and sexism. Furthermore, in the real world, hateful posts do not fall into to a single type of hate speech. There is a huge overlapping between different hateful classes, making hate speech identification a multi label problem. 

A wide variety of system architectures, ranging from classical machine learning to recent deep learning models, have been tried for various aspects of hate speech identification. Facebook, YouTube, and Twitter are the major sources of data for most data-sets.  \cite{Burnap} used SVM and ensemble techniques on identifying hate speech in Twitter data. \cite{Razavi2010} approach for abuse detection using an insulting and abusive language dictionary of words and phrases. \cite{Hee2015} used bag of words n-gram features and trained an SVM model on a cyberbullying dataset. \cite{Salminen2018} achieved an F1-score of 0.79 on classification of hateful YouTube and Facebook posts using a linear SVM model employing TF-IDF weighted n-grams. 

Recently, models based on deep learning techniques have also been applied to the task of hate speech identification. These models often rely on distributed representations or embeddings, e.g., FastText embeddings (\cite{joulin-etal-2017-bag}, and paragraph2vec distributed representations (\cite{DBLP:journals/corr/LeM14}. \cite{badjatya} employed an LSTM architecture to tune Glove word embeddings on the DATA-TWITTER-TWH data-set. \cite{risch2018aggression} used a neural network architecture using a GRU layer and ensemble methods for the TRAC 2018 \citep{TRAC2018} shared task on aggression identification. They also tried back-translation as a data augmentation technique to increase the data-set size. \cite{wang-2018-interpreting} illustrated the use of sequentially combining CNNs with RNNs  for abuse detection. They show that this approach was better than using only the CNN architecture giving a 1\% improvement in the F1-score.
One of the most recent developments in NLP are the transformer architecture introduced by \cite{vaswani2017attention}. Utilizing the transformer architecture, \cite{Bert} provide methods to pre-train models for language understanding (BERT) that have achieved state of the art results in many NLP tasks and are promising for hate speech detection as well. BERT based models achieved competitive performance in HASOC 2019 shared tasks. We \cite{Mishra2019HASOC} fine tuned BERT base models for the various HASOC shared tasks being the top performing model in some of the sub-tasks. A similar approach was also used for the TRAC 2020 shared tasks on Aggression identification by us \cite{Mishra2020TRAC} still achieving competitive performance with the other models without using an ensemble techniques. An interesting approach was the use of multi-lingual models by joint training on different languages. This approach presents us with a unified model for different languages in abusive content detection. Ensemble techniques using BERT models \citep{risch2020bagging} was the top performing model in many of the shared tasks in TRAC 2020. Recently, multi-task learning has been used for improving performance on NLP tasks \citep{Liu2016MTL,Sogaard2016}, especially social media information extraction tasks \citep{Mishra2019MDMT}, and more simpler variants have been tried for hate speech identification in our recent works \citep{Mishra2019HASOC,Mishra2020TRAC}. \cite{Florio_2020} investigated the usage of AlBERTo on monitoring hate speech against Italian on Twitter. Their results show that even though AlBERTo is sensitive to the fine tuning set, it's performance increases given enough training time. \cite{mozafari-marzieh} employ a transfer learning approach using BERT for hate speech detection. \cite{ranasinghe-zampieri-2020-multilingual} use cross-lingual embeddings to identify offensive content in multilingual setting. Our multi-lingual approach is similar in spirit to the method proposed in \cite{plank-2017-1} which use the same model architecture and aligned word embedding to solve the tasks.  There has also been some work on developing solutions for multilingual toxic comments which can be related to hate speech.\footnote{\url{https://www.kaggle.com/c/jigsaw-multilingual-toxic-comment-classification}} Recently, \cite{Mishra2020EDNIL} also used a single model across various tasks which performed very well for event detection tasks for five Indian languages. 

There have been numerous competitions dealing with hate speech evaluation. OffensEval \cite{zampieri-etal-2019-semeval} is one of the popular shared tasks dealing with offensive language in social media, featuring three sub-tasks for discriminating between offensive and non-offensive posts. Another popular shared task in SemEval is the HateEval \cite{Basile2019} task on the detection of hate against women and immigrants. The 2019 version of HateEval consists of two sub-task for determination of hateful and aggressive posts. GermEval \cite{germeval} is another shared task quite similar to HASOC. It focused on the Identification of Offensive Language in German Tweets. It features two sub-tasks following a binary and multi-class classification of the German tweets.

An important aspect of hate speech is that it is primarily multi-modal in nature. A large portion of the hateful content that is shared on social media is in the form of memes, which feature multiple modalities like audio, text, images and videos in some cases as well. \cite{yang-etal-2019-exploring-deep} present different fusion approaches to tackle multi-modal information for hate speech detection.  \cite{multimodal-hatespeech} explore multi-modal hate speech consisting of text and image modalities. They propose various multi-modal architectures to jointly analyze both the textual and visual information. Facebook recently released the hateful memes data-set for the Hateful Memes challenge \cite{kiela2020hateful} to provide a complex data-set where it is difficult for uni-modal models to achieve good performance.


\section{Methods}
\label{sec:methods}
For this paper, we extend some of the techniques that we have used in TRAC 2020 in \cite{Mishra2020TRAC} as well as \cite{Mishra2019MDMT,Mishra2020ACMSigIR,MishraThesisDSTDIE2020}, and apply them to the HASOC data-set \cite{HASOC2019}. Furthermore, we extend the work that we did as part of the HASOC 2019 shared task \cite{Mishra2019HASOC} by experimenting with multi-lingual training, back-translation based data-augmentation, and multi-task learning to tackle the data sparsity issue of the HASOC 2019 data-set. 

\begin{figure}[!htb]
    \centering
    \includegraphics[width=1.0\textwidth]{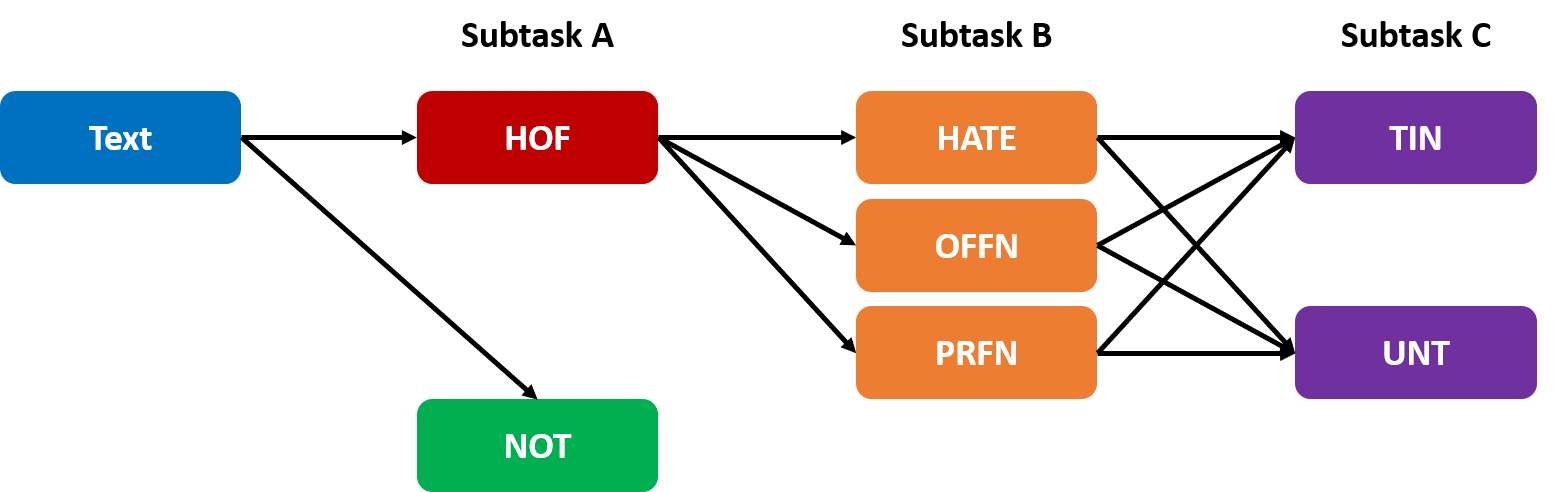}
    \caption{Task Description }
    \label{fig:task_desc}
\end{figure}

\subsection{Task Definition and Data}
All of the experiments reported hereafter have been done on the HASOC 2019 data-set \citep{HASOC2019} consisting of posts in English (EN), Hindi (HI) and German (DE). The shared tasks of HASOC 2019 had three sub-tasks \textbf{(A,B,C)} for both English and Hindi languages and two sub-tasks \textbf{(A,B)} for the German language. The description of each sub-task is as follows (see Figure \ref{fig:task_desc} for details):
\begin{itemize}
    \item \textbf{Sub-Task A} : Posts have to be classified into hate speech \textbf{HOF} and non-offensive content \textbf{NOT}.
    \item \textbf{Sub-Task B} : A fine grained classification of the hateful posts in sub-task A. Hate Speech posts have to be identified into the type of hate they represent, i.e containing hate speech content \textbf{(HATE)}, containing offensive content \textbf{(OFFN)} and those containing profane words \textbf{(PRFN)}.
    \item \textbf{Sub-Task C} : Another fine grained classification of the hateful posts in sub-tasks A. This sub-task required us to identify whether the hate speech was targeted towards an individual or group \textbf{TIN} or whether it was un-targeted \textbf{UNT}.
\end{itemize}

\begin{table}[!htb]
    \centering
    \caption{Distribution of number of tweets in different data-sets and splits.}
    \label{tab:data_distribution}
    \begin{tabular}{l|rrr|rrr|rrr}
    \toprule
    \textbf{task} & \multicolumn{3}{c}{\textbf{DE}} & \multicolumn{3}{c}{\textbf{EN}} & \multicolumn{3}{c}{\textbf{HI}} \\
    &  train &  dev &  test &  train &  dev &  test &  train &  dev &  test\\
    \midrule
    \textbf{A} &   3,819 &  794 &   850 &   5,852 &  505 &  1,153 &   4,665 &  136 &  1,318\\
    \textbf{B} &    407 &  794 &   850 &   2,261 &  302 &  1,153 &   2,469 &  136 &  1,318 \\
    \textbf{C} &        &      &       &   2,261 &  299 &  1,153 &   2,469 &   72 &  1,318 \\
    \bottomrule
    \end{tabular}
\end{table}

The HASOC 2019 data-set consists of posts taken from Twitter and Facebook. The data-set only consists of text and labels and does not include any contextual information or meta-data of the original post e.g. time information. The data distribution for each language and sub-task is mentioned in Table \ref{tab:data_distribution}. We can observe, that the sample size for each language is of the order of a few thousand post, which is an order smaller to other datasets like OffenseEval (13,200 posts), HateEval (19,000 posts), and Kaggle Toxic Comments datasets (240,000 posts). This can pose a challenge for training deep learning models, which often consists of large number of parameters, from scratch. Class wise data distribution for each language is available in the appendix \ref{appendix:label_dist} figures \ref{fig:data_dist:a}, \ref{fig:data_dist:b}, and \ref{fig:data_dist:c}. These figures show that the label distribution is highly skewed for task C,  such as the label UNT, which is quite underrepresented. Similarly, for German the task A data is quite unbalanced. For more details on the dataset along with the details on its creation and motivation we refer the reader to \cite{HASOC2019}. \cite{HASOC2019} reports that the inter-annotator agreement is in the range of 60\% to 70\% for English and Hindi. Furthermore, the inter-annotator agreement is more than 86\% for German.  

\begin{figure}
    \centering
    \includegraphics[width=\textwidth]{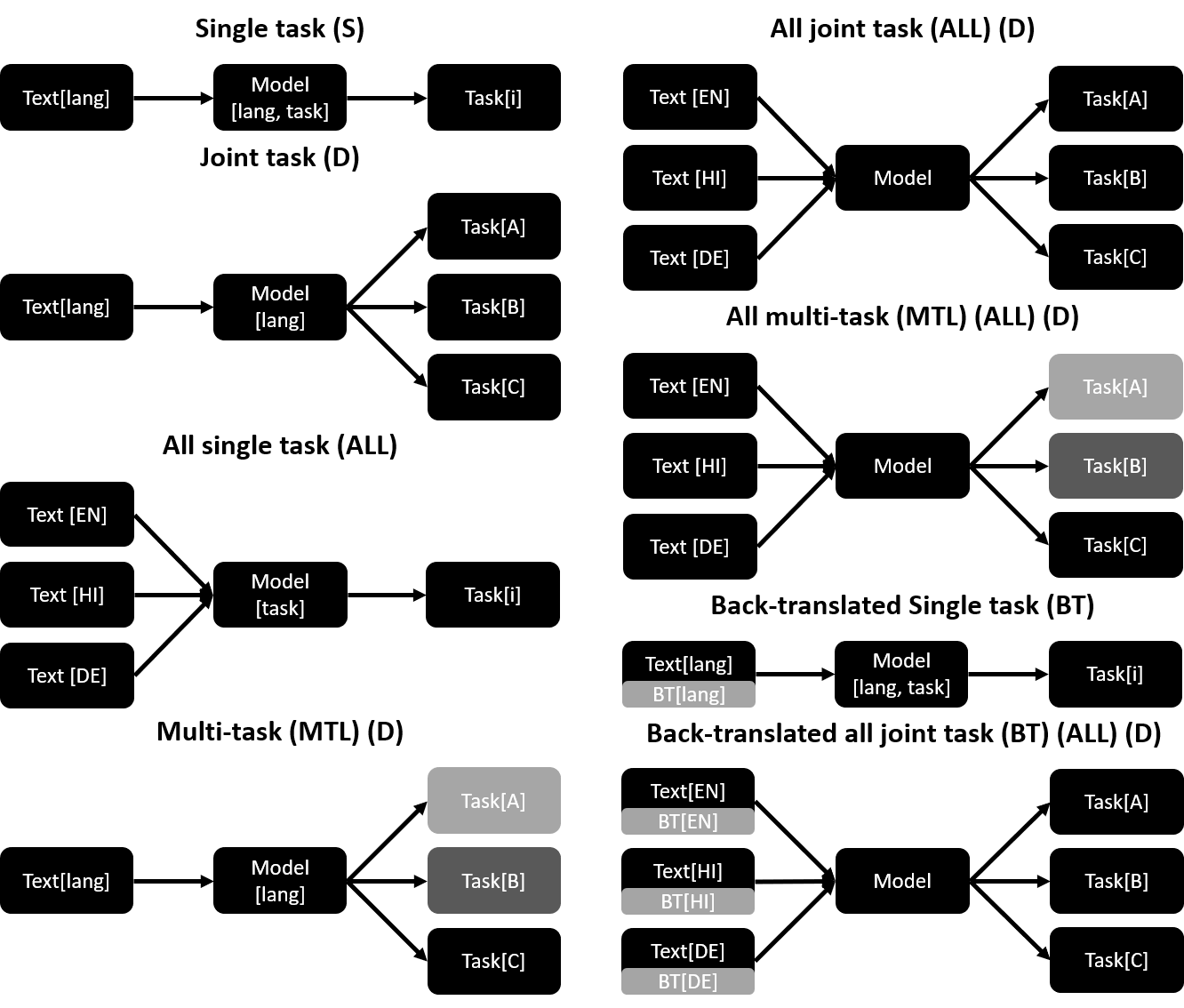}
    \caption{An overview of various model architectures we used. Shaded task boxes represent that we first compute a marginal representation of labels only belonging to that task before computing the loss.}
    \label{fig:models}
\end{figure}

\subsection{Fine-tuning transformer based models}

The transformer based models especially BERT \citep{Bert}, have proven to be successful in achieving very good results on a range of NLP tasks. Upon its release, BERT based models became state of the art for 11 NLP tasks \citep{Bert}. This motivated us to try out BERT for hate speech detection. We had used multiple variants of the BERT model during HASOC 2019 shared tasks \cite{Mishra2019HASOC}. We also experimented with other transformer models and BERT during TRAC2020 \cite{Mishra2020TRAC}. However, based on our experiments, we find the original BERT models to be best performing for most tasks. Hence, for this paper we only implement our models on those. For our experiments we use the open source implementations of BERT provided by  \cite{wolf2019huggingfaces}\footnote{\href{https://github.com/huggingface/transformers}{https://github.com/huggingface/transformers}}. A common practice for using BERT based models, is to fine-tune an existing pre-trained model on data from a new task. For fine tuning the pre-trained BERT models we used the BERT for Sequence Classification paradigm present in the HuggingFace library. We fine tune BERT using various architectures. A visual description of these architectures is shown in Figure \ref{fig:models}. These models are explained in detail in later sections. 

To process the text, we first use a pre-trained BERT tokenizer to convert the input sentences into tokens. These tokens are then passed to the model which generate a BERT specific embeddings for each token. The special part about BERT is that its decoder is supplied all the hidden states of the encoder unlike other transformer models before BERT. This helps it to capture better contextual information even for larger sequences. Each sequence of tokens is padded with a [CLS] and [SEP] token. The pre-trained BERT model generates an output vector for each of the tokens. For sequence classification tasks, the vector corresponding to the [CLS] token is used as it holds the contextual information about the complete sentence. Additional fine-tuning is done on this vector to generate the classification for specific data-sets. 

To keep our experiments consistent, the following hyper-parameters were kept constant for all our experiments. For training our models we used the standard hyper-parameters as mentioned in the huggingface transformers documentation. We used the Adam optimizer (with $\epsilon = 1e-8$) for 5 epochs, with a training/eval batch size of 32. Maximum allowable length for each sequence was kept as $128$. We use a linearly decreasing learning rate with a starting value as $5e-5$ with a weight decay of 0.0 and a max gradient norm of $1.0$. All models were trained using Google Colab's\footnote{\href{https://colab.research.google.com/}{https://colab.research.google.com/}} GPU runtimes. This limited us to a model run-time of 12 hours with a GPU which constrained our batch size as well as number of training epochs based on the GPU allocated by Colab. 

We refer to models which fine tune BERT on using data set from a single language for a single task, as \emph{Single models} with an indicator \emph{(S)}, this is depicted in Figure \ref{fig:models} (1st row left). All other models types which we discuss later are identified by their model types and naems in Figure \ref{fig:models}.

\subsection{Training a model for all tasks}
One of the techniques that we had used for our work in HASOC 2019 \cite{Mishra2019HASOC} was creating an additional sub-task D by combining the labels for all of the sub-tasks. We refer to models which use this technique as \emph{Joint task models} which an indicator \emph{(D)} (see Figure \ref{fig:models} models marked with D). This allowed us to train a single model for all of the sub-tasks. This also helps in overcoming the data sparsity issue for sub-tasks B and sub-tasks C for which the no. of data points is very small. The same technique was also employed in our submission to TRAC 2020 \cite{Mishra2020TRAC} aggression and misogyny identification tasks. Furthermore, when combining labels, we only consider valid combination of labels, which allows us to reduce the possible output space. For HASOC, the predicted output labels for the joint-training are as follows : \textbf{NOT-NONE-NONE}, \textbf{HOF-HATE-TIN}, \textbf{HOF-HATE-UNT}, \textbf{HOF-OFFN-TIN}, 
\textbf{HOF-OFFN-UNT}, \textbf{HOF-PRFN-TIN}, \textbf{HOF-PRFN-UNT}. The task specific labels can be easily extracted from the output labels, using post-processing of predicted labels. 



\subsection{Multi-lingual training}
Inspired from joint training of all tasks, as described above, we also implement the training of a single model for all languages for a given sub-task. Similar approach was 
utilized in our prior submission to TRAC 2020 \cite{Mishra2020TRAC}. We refer to models which use this technique as \emph{All models} with an indicator \emph{(ALL)} (see Figure \ref{fig:models} models marked with ALL). In this method, we combine the data-sets from all the languages and train a single multi-lingual model on this combined data-set. The multi-lingual model is able to learn data from multiple languages thus providing us with a single unified model for different languages. A major motivation for taking this approach was that social media data often does not belong to one particular language. It is quite common to find code-mixed posts on Twitter and Facebook. Thus, a multi-lingual model is the best choice in this scenario. During our TRAC 2020 work, we had found out that this approach works really well and was one of our top models in almost all of the shared tasks. From a deep learning point of view, this technique seems promising as doing this also increases the size of the data-set available for training without adding new data points from other data-sets or from data augmentation techniques. 

As a natural extension of the above two approaches, we combine the multi-lingual training with the joint training approach to train a single model on all tasks for all languages. We refer to models which use this technique as \emph{All joint task models} with an indicator \emph{(ALL) (D)} (see Figure \ref{fig:models}).


\subsection{Multi-task learning}

While the joint task setting, can be considered as a multi-task setting, it is not, in the common sense, and hence our reservation in calling it multi-task. The joint task training can be considered an instance of multi-class prediction, where the number of classes is based on the combination of tasks. This approach does not impose any sort of task specific structure on the model, or computes and combines task specific losses. The core idea of multi-task learning is to use similar tasks as regularizers for the model. This is done by simply adding the loss functions specific to each task in the final loss function of the model. This way the model is forced to optimize for all of the different tasks simultaneously, thus producing a model that is able to generalize on multiple tasks on the data-set. However, this may not always prove to be beneficial as it has been reported that when the tasks differ significantly the model fails to optimize on any of the tasks. This leads to significantly worse performance compared to single task approaches. However, sub-tasks in hate speech detection are often similar or overlapping in nature. Thus, this approach seems promising for hate speech detection.

Our multi-task setup is inspired from the marginalized inference technique which was used in \cite{Mishra2020TRAC}. In the marginalized inference, we post-process the probabilities of each label in the joint model, and compute the task specific label probability by marginalizing the probability of across all the other tasks. This ensures that the probabilities of labels for each sub-task make a valid probability distribution and sum to one. For example, $p(\textbf{HOF-HATE-TIN}) > p(\textbf{HOF-PRFN-TIN})$ does not guarantee that $p(\textbf{HOF-HATE-UNT}) > p(\textbf{HOF-PRFN-UNT})$. As described above, we can calculate the task specific probabilities by marginalizing the output probabilities of that task. For example, $p(\textbf{HATE}) = p(\textbf{HOF-HATE-TIN}) + p(\textbf{HOF-HATE-UNT})$. However, using this technique did not lead to a significant improvement in the predictions and the evaluation performance. In some cases, it was even lower than the original method. A reason we suspect for this low performance is that the model was not trained to directly optimize its loss for this marginal inference. Next, we describe our multi-task setup inspired from this approach. 

For our multi-task experiments, we first use our joint training approach (sub-task D) to generate the logits for the different class labels. These logits are then marginalized to generate task specific logits (marginalizing logits is simpler than marginalizing the probability for each label, as we do not need to compute the partition function). For each task, we take a cross-entropy loss using the new task specific logits. Finally we add the respective losses for each sub-task along with the sub-task D loss. This added up loss is the final multi-task loss function of our model. We then train our model to minimize this loss function. In this loss, each sub-task loss acts as a regularizer for the other task losses. Since, we are computing the multi-task loss for each instance, we include a special label \emph{NONE} for sub-tasks B and C, for the cases where the label of sub-task A is \emph{NOT}. We refer to models which use this technique as \emph{Multi-task models} with an indicator \emph{(MTL) (D)} (see Figure \ref{fig:models}).

One important point to note is that we restrict the output space of the multi-task model by using the task 4 labels. This is an essential constraint that we put on the model because of which there is no chance of any inconsistency in the prediction. By inconsistency we say that it is not possible for our multi-task model to predict a data point that belongs to \emph{NOT} for task A and to any label other than \emph{NONE} for the tasks B and C. If we follow the general procedure for training a multi-task model, we would have \textbf{$2 * (3+1) * (2+1) = 24$} , with $+1$ for the additional \emph{NONE} label, combinations of outputs from our model, which would produce the inconsistencies mentioned above. 

Like the methods mentioned before, we extend multi-task learning to all languages, which results in \emph{Multi-task all model}, which are indicated with an indicator \emph{(MTL) (ALL) (D)}. 


\subsection{Training with Back-Translated data}
\label{sec:methods:backtranslate}
One approach for increasing the size of the training data-set, is to generate new instances based on existing instances using data augmentation techniques. These new instances are assigned the same label as the original instance. Training model with instances generated using data augmentation techniques assumes that the label remains same if the data augmentation does not change the instance significantly. We utilized a specific data augmentation technique used in NLP models, called Back-Translation \citep{Koehn2005,Sennrich2016}. Back-translation uses two machine translation models, one, to translate a text from its original language to a target language; and another, to translate the new text in target language back to the original language. This technique was successfully used in the submission of \cite{risch2018aggression,risch2020bagging} during TRAC 2018 and 2020.  
Data augmentation via back-translation assumes that current machine translation systems when used in back-translation settings give a different text which expresses a similar meaning as the original. This assumption allows us to reuse the label of the original text for the back-translated text. 

We used the Google translate API\footnote{\href{https://cloud.google.com/translate/docs}{https://cloud.google.com/translate/docs}} to back-translate all the text in our data-sets.

For each language in our data-set we use the following source $\rightarrow$ target $\rightarrow$ source pairs:  

\begin{itemize}
    \item \textbf{EN}: \emph{English} $\rightarrow$ \emph{French}  $\rightarrow$ \emph{English}
    \item \textbf{HI}: \emph{Hindi}   $\rightarrow$ \emph{English} $\rightarrow$ \emph{Hindi}
    \item \textbf{DE}: \emph{German}  $\rightarrow$ \emph{English} $\rightarrow$ \emph{German}
\end{itemize}

To keep track of the back-translated texts we added a flag to the text id. In many cases, there were minimal changes to the text. In some cases there were no changes to the back-translated texts. However the no. of such texts where there was no change after back-translation was very low. For example, among 4000 instances in the English training set around 100 instances did not have any changes. So while using the back-translated texts for our experiments, we simply used all the back-translated texts whether they under-went a change or not. The data-set size doubled after using the back-translation data augmentation technique. An example of back-translated English text is as follows (changed text is emphasized):
\begin{enumerate}
    \item \textbf{Original}: {@politico No. We should remember very clearly that \#Individual1 just admitted to treason . \#TrumpIsATraitor  \#McCainsAHero \#JohnMcCainDay}
    \item \textbf{Back-translated}: {@politico No, \textbf{we must not forget that very clear} \#Individual1 just admitted to treason. \#TrumpIsATraitor \#McCainsAHero \#JohnMcCainDay}
\end{enumerate}

\section{Results}
\label{sec:results}
We present our results for sub-tasks A, B and C in Table \ref{tab:A_results}, \ref{tab:B_results}, and \ref{tab:C_results} respectively. To keep the table concise we use the following convention. 

\begin{enumerate}[align=left]
    \item \textbf{(ALL)}: A \emph{bert-base-multi-lingual-uncased} model was used with multi-lingual joint training.
    \item \textbf{(BT)}: The data-set used for this experiment is augmented using back-translation.
    \item \textbf{(D)}: A joint training approach has been used.
    \item \textbf{(MTL)}: The experiment is performed using a multi-task learning approach.
    \item \textbf{(S)}: This is the best model which was submitted to HASOC 2019 in \cite{Mishra2019HASOC}.
\end{enumerate}

The pre-trained BERT models which were fine-tuned for each language in a single language setting, are as follows:
 \begin{enumerate}
     \item \textbf{EN} - \emph{bert-base-uncased}
     \item \textbf{HI} - \emph{bert-base-multi-lingual-uncased}
     \item \textbf{DE} - \emph{bert-base-multi-lingual-uncased}
 \end{enumerate}



\newcommand{\bftab}{\fontseries{b}\selectfont}

\subsection{Model performance}
We evaluate our models against each other and also against the top performing models of HASOC 2019 for each task.
We use the same benchmark scores, namely, \textbf{weighted F1-score} and \textbf{macro F1-score}, as were used in \cite{HASOC2019}, with \textbf{macro F1-score} being the scores which were used for overall ranking in HASOC 2019.

\subsubsection{Sub-Task A}

\begin{table}[]
    \centering
    \caption{Sub-task A results. Models in HASOC 2019 \citep{HASOC2019} were ranked based on Macro F1.}
    \label{tab:A_results}
    \begin{tabular}{ll|rrr|rrr}
\toprule
      & {} & \multicolumn{3}{c}{Weighted F1} & \multicolumn{3}{c}{Macro F1} \\
   \textbf{lang} & \textbf{model} &          dev &  train &   test &       dev &  train &   test \\
\midrule
\multirow{7}{*}{\textbf{EN}} & \textbf{(ALL) } &        0.562 &  0.949 &  0.804 &     0.568 &  0.946 &  0.753 \\
   & \textbf{(ALL) (D) } &        0.481 &  0.894 &  0.797 &     0.497 &  0.886 &  0.740 \\
   & \textbf{(BT) } &        0.535 &  0.973 &  0.756 &     0.545 &  0.971 &  0.690 \\
   & \textbf{(BT) (ALL) } &        0.493 &  0.986 &  0.803 &     0.509 &  0.985 &  0.747 \\
   & \textbf{(BT) (ALL) (D) } &        0.474 &  0.981 &  0.806 &     0.492 &  0.980 &  0.750 \\
   & \textbf{(MTL) (ALL) (D) } &        0.552 &  0.823 &  0.801 &     0.559 &  0.812 &  0.755 \\
   & \textbf{(MTL) (D) } &        0.543 &  0.745 &  \bftab 0.819 &     0.557 &  0.725 &  \bftab 0.765 \\
   \cmidrule{2-8}
 &  \textbf{(S)}  & 0.606  & 0.966  & {0.790} & 0.610  & 0.964 & {0.740} \\
 &  \textbf{(S) (D)}  & 0.596  & 0.908  & {0.801} & 0.603  & 0.902 & {0.747} \\
 &  \textbf{HASOC Best}  & -  & -  & \bftab{0.840} & -  & - & \bftab{0.788} \\
\midrule
\multirow{7}{*}{\textbf{HI}} & \textbf{(ALL) } &        0.786 &  0.976 &  0.793 &     0.785 &  0.976 &  0.793 \\
   & \textbf{(ALL) (D) } &        0.815 &  0.959 &  0.811 &     0.815 &  0.959 &  0.810 \\
   & \textbf{(BT) } &        0.654 &  0.967 &  0.746 &     0.654 &  0.967 &  0.744 \\
   & \textbf{(BT) (ALL) } &        0.815 &  0.982 &  0.795 &     0.814 &  0.982 &  0.795 \\
   & \textbf{(BT) (ALL) (D) } &        0.772 &  0.975 &  0.793 &     0.772 &  0.975 &  0.792 \\
   & \textbf{(MTL) (ALL) (D) } &        0.860 &  0.921 &  0.808 &     0.860 &  0.921 &  0.807 \\
   & \textbf{(MTL) (D) } &        0.748 &  0.893 &  \bftab 0.814 &     0.749 &  0.893 &  \bftab 0.814 \\
   \cmidrule{2-8}
 &  \textbf{(S)}  & 0.742 & 0.961  & {0.802} & 0.742  & 0.961 & {0.802} \\
 &  \textbf{(S) (D)}  &  0.822 & 0.941  & {0.814} & 0.823 & 0.941 & {0.811} \\
 &  \textbf{HASOC Best}  & -  &  - & \textbf{0.820} &  - & - & \textbf{0.815} \\
\midrule
\multirow{7}{*}{\textbf{DE}} & \textbf{(ALL) } &        0.899 &  0.993 &  0.794 &     0.706 &  0.981 &  0.584 \\
   & \textbf{(ALL) (D) } &        0.906 &  0.988 &  0.779 &     0.730 &  0.968 &  0.566 \\
   & \textbf{(BT) } &        0.878 &  0.988 &  0.777 &     0.628 &  0.969 &  0.533 \\
   & \textbf{(BT) (ALL) } &        0.908 &  0.999 &  \bftab 0.800 &     0.742 &  0.998 &  \bftab 0.612 \\
   & \textbf{(BT) (ALL) (D) } &        0.902 &  0.998 &  0.783 &     0.712 &  0.994 &  0.584 \\
   & \textbf{(MTL) (ALL) (D) } &        0.917 &  0.969 &  0.786 &     0.764 &  0.915 &  0.582 \\
   & \textbf{(MTL) (D) } &        0.878 &  0.898 &  0.789 &     0.593 &  0.683 &  0.526 \\
   \cmidrule{2-8}
 &  \textbf{(S)}  &  0.606 & 0.966  & {0.789} & 0.610  & 0.964 & {0.577} \\
 &  \textbf{HASOC Best}  &  - & -  & \bftab{0.792} & -  & - & \bftab{0.616} \\
\bottomrule
\end{tabular}
\end{table}

The best scores for sub-task A are mentioned in Table \ref{tab:A_results}. The best scores for this task belong to \cite{wang2019ynu_wb}, \cite{bashar2020qutnocturnal} and \cite{hatemonitors} for English, Hindi and German respectively.
All the models that we experimented with in sub-task A are very closely separated by the macro-F1 score. Hence, all of them give a similar performance for this task. The difference between the macro F1-scores of these models is $< 3\%$ . For both English and Hindi, the multi-task learning model performed the best while for the German language, the model that was trained on the back-translated data using the multi-lingual joint training approach and task D \textbf{( (ALL) (BT) (D) )} worked best. However, it is interesting to see that the multi-task model gives competitive performance on all of the languages within the same computation budget. One thing to notice is that, the train macro-F1 scores of the multi-task model are much lower than the other models. This suggests that the \textbf{(MTL)} model, given additional training time might improve the results even further. We are unable to provide longer training time due to lack of computational resources available to us. The \textbf{(ALL) (MTL)} model also gives a similar performance compared to the (MTL) model. This suggests that the additional multi-lingual training comes with a trade off with a slightly lower macro-F1 score.
However, the difference between the scores of the two models is $\sim 1\%$. In order to address the additional training time the (MTL) models required, we trained the (ALL) (MTL) model for 15 epochs. However, this training time was too large as the models over-fitted the data. This finally resulted in a degradation in the performance of these models. A sweet spot for the training time may be found for the (MTL) models which may result in an increase in the performance of the model whilst avoiding over-fitting. We were not able to conduct more experiments to do the same due to time constraints. This may be evaluated in the additional future work on these models. We, however, cannot compare the German (MTL) models with the (MTL) models of the other languages as the German data did not have not have sub-task C, so the (MTL) approach did not have sub-task C for German. As we will see in the next section, the (MTL) models performed equally well in sub-task B. This might be because both tasks A and B involve identifying hate and hence are in a sense co-related. This co-relation is something that the (MTL) models can utilize for their advantage. It has been found in other multi-task approaches that the models learn more effectively when the different tasks are co-related. However, their performance can degrade if the tasks are un-related. The lower performance on the German data may be because of the unavailability of the sub-task C. However, the results are still competitive with the other models. For German, the (ALL) (MTL) model performed better than our submission for HASOC 2019. The (MTL) model for Hindi was able to match the best model for this task at HASOC 2019. 

The (ALL) and (ALL) (D) training methods show an improvement from our single models submitted at HASOC. These models present us with an interesting option for abuse detection tasks as they are able to work on all of the shared tasks at the same time, leveraging the multi-lingual abilities of the model whilst still having a computation budget equivalent to that of a single model. These results show that these models give a competitive performance with the single models. They even outperform the single model, e.g., they outperform the \textbf{bert-base-uncased} single models that were used in English sub-task A, which have been specially tuned for English tasks. While for German and Hindi, the single models themselves utilized a bert-base-uncased model, so they are better suited for analyzing the improvements by the multi-lingual joint training approach. On these languages we see, that the (ALL) and (ALL) (D) techniques do improve the macro-F1 scores on for this task. 

The back-translation technique does not seem to improve the models much. The approach had a mixed performance. For all the languages, back-translation alone does not improve the model and hints at over-fitting, resulting in a decrease in test results. However, when it is combined with the (ALL) and (D) training methods we see an increase in the performance. The (ALL) and (D) training methods are able to leverage the data-augmentation applied in the back-translated data. Back-translation when used with (ALL) or (ALL) (D) are better than the single models that we submitted at HASOC 2019. The (BT) (ALL) model comes really close to the best model at HASOC, coming second according to the results in \cite{HASOC2019}. 

\subsubsection{Sub-Task B}

\begin{table}[]
    \centering
    \caption{sub-task B results. Models in HASOC 2019 \citep{HASOC2019} were ranked based on Macro F1.}
    \label{tab:B_results}
    \begin{tabular}{ll|rrr|rrr}
\toprule
      & {} & \multicolumn{3}{c}{Weighted F1} & \multicolumn{3}{c}{Macro F1} \\
   \textbf{lang} & \textbf{model} &          dev &  train &   test &       dev &  train &   test \\
\midrule
\multirow{7}{*}{\textbf{EN}} & \textbf{(ALL) } &        0.361 &  0.826 &  0.501 &     0.290 &  0.805 &  0.467 \\
   & \textbf{(ALL) (D) } &        0.201 &  0.776 &  0.556 &     0.190 &  0.580 &  0.392 \\
   & \textbf{(BT) } &        0.422 &  0.965 &  0.532 &     0.352 &  0.960 &  0.510 \\
   & \textbf{(BT) (ALL) } &        0.396 &  0.962 &  0.626 &     0.375 &  0.957 &  0.591 \\
   & \textbf{(BT) (ALL) (D) } &        0.201 &  0.950 &  0.576 &     0.153 &  0.708 &  0.408 \\
   & \textbf{(MTL) (ALL) (D) } &        0.397 &  0.927 &  0.635 &     0.277 &  0.915 &  0.590 \\
   & \textbf{(MTL) (D) } &        0.344 &  0.899 &  \bftab 0.638 &     0.341 &  0.881 &  \bftab 0.600 \\
   \cmidrule{2-8}
 &  \textbf{(S)}  & 0.349  & 0.867  & {0.728} & 0.314  & 0.846 & {0.545} \\
 &  \textbf{(S) (D)}  & 0.401  & 0.875  & {0.698} & 0.332  & 0.839 & {0.537} \\
 &  \textbf{HASOC Best}  & -  & -  & \bftab{0.728} & -  & - & \bftab{0.545} \\
\midrule
\multirow{7}{*}{\textbf{HI}} & \textbf{(ALL) } &        0.494 &  0.832 &  0.500 &     0.340 &  0.802 &  0.494 \\
   & \textbf{(ALL) (D) } &        0.678 &  0.792 &  0.564 &     0.293 &  0.566 &  0.415 \\
   & \textbf{(BT) } &        0.231 &  0.807 &  0.507 &     0.160 &  0.767 &  0.501 \\
   & \textbf{(BT) (ALL) } &        0.589 &  0.890 &  \bftab 0.667 &     0.283 &  0.875 &  \bftab 0.662 \\
   & \textbf{(BT) (ALL) (D) } &        0.630 &  0.849 &  0.519 &     0.180 &  0.617 &  0.381 \\
   & \textbf{(MTL) (ALL) (D) } &        0.819 &  0.883 &  0.647 &     0.499 &  0.864 &  0.641 \\
   & \textbf{(MTL) (D) } &        0.553 &  0.802 &  0.602 &     0.348 &  0.764 &  0.593 \\
   \cmidrule{2-8}
 &  \textbf{(S)}  & 0.466  & 0.749  & {0.688} & 0.322  & 0.701 & {0.553} \\
 &  \textbf{(S) (D)}  & 0.757   & 0.826  & {0.715} & 0.459  & 0.736 & {0.581} \\
 &  \textbf{HASOC Best}  & -  & -  & \bftab{0.715} & -  & - & \bftab{0.581} \\
\midrule
\multirow{7}{*}{\textbf{DE}} & \textbf{(ALL) } &        0.326 &  0.876 &  0.459 &     0.315 &  0.861 &  0.343 \\
   & \textbf{(ALL) (D) } &        0.285 &  0.813 &  0.154 &     0.255 &  0.603 &  0.128 \\
   & \textbf{(BT) } &        0.285 &  0.620 &  0.413 &     0.328 &  0.581 &  0.285 \\
   & \textbf{(BT) (ALL) } &        0.478 &  0.985 &  0.495 &     0.484 &  0.984 &  0.397 \\
   & \textbf{(BT) (ALL) (D) } &        0.179 &  0.945 &  0.242 &     0.153 &  0.707 &  0.177 \\
   & \textbf{(MTL) (ALL) (D) } &        0.463 &  0.946 &  0.527 &     0.346 &  0.707 &  0.344 \\
   & \textbf{(MTL) (D) } &        0.468 &  0.923 &  \bftab 0.541 &     0.482 &  0.918 &  \bftab 0.416 \\
\cmidrule{2-8}
 &  \textbf{(S)}  & 0.112   & 0.367  & {0.756} & 0.140    & 0.247 & {0.249} \\
 &  \textbf{(S) (D)}  &  0.865 & 0.918  & {0.778} & 0.282 & 0.409 & {0.276} \\
 &  \textbf{HASOC Best}  & -  & -  & \bftab{0.775} & -  & - & \bftab{0.347} \\
\bottomrule
\end{tabular}
\end{table}

The best scores for sub-task B are mentioned in Table \ref{tab:B_results}. The best scores for this task belong to \cite{DanaRuiter} for German. For English and Hindi sub-task B our submissions had performed the best at HASOC 2019.
For sub-task B, many of our models were able to significantly outperform the best HASOC models. 
For English, the multi-task approach results in a new best macro-F1 score of $0.600$, a $6\%$ increase from the previous best. For Hindi, our (BT) (ALL) results in a macro-F1 score of $0.662$ which is $8\%$ more than the previous best. For Germans, our (MTL) model has a macro-F1 score on the test set of 0.416 which is almost $7\%$ more than the previous best.

For the English task, even our (MTL) (ALL) and (BT) (ALL) models were able to beat the previous best. However, our results show that unlike sub-task A where our models had similar performances, in sub-task B there is huge variation in their performance. Many outperform the best, however some of them also show poor results. The (ALL) and (ALL) (D) perform poorly for the three languages, except (ALL) in German, and show very small macro-F1 scores even on the training set. Thus, training these models for longer may change the results. The (MTL) models are able to give competitive performance in task-A and is able to outperform the previous best, thus showing it's capability to leverage different co-related tasks and generalize well on all of them. 

Here again we see that back-translation alone does not improve the macro-F1 scores. However, an interesting thing to notice is that the (ALL) and (BT) models which perform poorly individually, tend to give good results when used together. Outperforming the previous best HASOC models, in all the three languages. This hints that data sparsity alone is not the major issue of this task. This is also evident from the performance of the (MTL) model which only utilizes the data-set of a single language, which is significantly smaller than the back-translated (twice the original data-set) and the multi-lingual joint model (sum of the sizes of the original model).  
But the (BT) (ALL) (D) model performed poorly in all of the three languages. Thus, the use of sub-task (D) with these models only degrades performance.

The results from this task confirm that the information required to predict task - A is important for task - B as well. This information is shared better by the modified loss function of the (MTL) models rather than the loss function for sub-task (D). 

The (MTL) models build up on the sub-task (D) approach and do not utilize it explicitly. The sub-task (D) approach seems like a multi-task learning method, however, it is not complete and is not able to learn from the other tasks, thus does not offer huge improvement. These (MTL) models do show a variation in their performance but it is always on the higher side of the macro-F1 scores.

\subsubsection{Sub-Task C}
The best scores for sub-task C are mentioned in Table \ref{tab:C_results}. The best scores for this task belong to \cite{mujadia2019iiit} for Hindi while our submissions performed the best for English.
The results for sub-task C also show appreciable variation. Except the (ALL) (D) and (BT) (ALL) (D) models which also performed poorly in sub-task B, the variation in their performance, especially for English, is not as significant as that present in sub-task B. This may be due to the fact that the two way fine-grained classification is a much easier task than the three way classification in sub-task B.
One important point to note is that sub-task C focused on identifying the context of the hate speech, specifically it focused on finding out whether it is targeted or un-targeted, while sub-task A and sub-task B both focused on identifying the type of hate speech. 

The (MTL) models do not perform as well as they did in the previous two tasks. They were still able to outperform the best models for English but perform poorly for Hindi. An important point to notice here is that the train macro-F1 scores for the (MTL) models is significantly low. This suggests that the (MTL) model was not able learn well even for the training instances of this task. This can be attributed to the point mentioned above that this task is inherently not as co-related to sub-task A and sub-task B as previously assumed. The task structure itself is not beneficial for a (MTL) approach. The main reason for this is that this task focuses on identifying targeted and un-targeted hate-speech. However, a non hate-speech text can also be an un-targeted or targeted text. As the (MTL) model receives texts belonging to both hate \textbf{(HOF)} and non-hate speech \textbf{NOT}, the information contained in the texts belonging to this task are counter acted by those targeted and un-targeted texts belong to the (NOT) class. Thus, a better description of this task is not a fine-grain classification of hate speech text, but that which involves targeted and un-targeted labels for both (HOF) and (NOT) classes. In that setting, we can fully utilize the advantage of the multi-task learning model and can expect better performance on this task as well.

The (ALL) and (BT) models performed really well in sub-task C. The (ALL), (BT) and (ALL) (BT) models outperform the previous best for English. The combination of these models with (D) still does not improve them and they continue to give poor performance. This provides more evidence to our previous inference that sub-task (D) alone does not improve the our performance.

\begin{table}[]
    \centering
    \caption{sub-task C results. Models in HASOC 2019 \citep{HASOC2019} were ranked based on Macro F1.}
    \label{tab:C_results}
    \begin{tabular}{ll|rrr|rrr}
\toprule
   & {} & \multicolumn{3}{c}{Weighted F1} & \multicolumn{3}{c}{Macro F1} \\
   \textbf{lang} & \textbf{model} &          dev &  train &   test &       dev &  train &   test \\
\midrule
\multirow{7}{*}{\textbf{EN}} & \textbf{(ALL) } &        0.842 &  0.986 &  0.737 &     0.524 &  0.958 &  \bftab 0.547 \\
   & \textbf{(ALL) (D) } &        0.380 &  0.792 &  0.658 &     0.141 &  0.328 &  0.292 \\
   & \textbf{(BT) } &        0.836 &  0.991 &  \bftab 0.771 &     0.465 &  0.974 &  0.543 \\
   & \textbf{(BT) (ALL) } &        0.839 &  0.987 &  0.718 &     0.534 &  0.962 &  0.514 \\
   & \textbf{(BT) (ALL) (D) } &        0.374 &  0.967 &  0.600 &     0.173 &  0.618 &  0.297 \\
   & \textbf{(MTL) (ALL) (D) } &        0.839 &  0.704 &  0.658 &     0.311 &  0.465 &  0.518 \\
   & \textbf{(MTL) (D) } &        0.844 &  0.747 &  0.692 &     0.470 &  0.506 &  0.538 \\
   \cmidrule{2-8}
 &  \textbf{(S)}  & 0.880  & 0.980 & {0.756} & 0.627  & 0.942 & {0.511} \\
 &  \textbf{(S) (D)}  &  0.548   & 0.874  & {0.764} & 0.393 & 0.651 & {0.476} \\
 &  \textbf{HASOC Best}  & -  & -  & \bftab{0.756} & -  & - & \bftab{0.511} \\
\midrule
\multirow{7}{*}{\textbf{HI}} & \textbf{(ALL) } &        0.844 &  0.765 &  \bftab 0.827 &     0.525 &  0.740 &  0.557 \\
   & \textbf{(ALL) (D) } &        0.594 &  0.666 &  0.740 &     0.216 &  0.417 &  0.336 \\
   & \textbf{(BT) } &        0.861 &  0.766 &  0.817 &     0.652 &  0.744 &  \bftab 0.568 \\
   & \textbf{(BT) (ALL) } &        0.797 &  0.941 &  0.775 &     0.517 &  0.937 &  0.527 \\
   & \textbf{(BT) (ALL) (D) } &        0.682 &  0.779 &  0.673 &     0.288 &  0.507 &  0.317 \\
   & \textbf{(MTL) (ALL) (D) } &        0.374 &  0.530 &  0.626 &     0.292 &  0.524 &  0.456 \\
   & \textbf{(MTL) (D) } &        0.557 &  0.577 &  0.628 &     0.397 &  0.573 &  0.451 \\
 \cmidrule{2-8}
  &  \textbf{(S)}  & 0.800 & 0.877 & {0.727} & 0.550  & 0.866 & {0.565} \\
 &  \textbf{(S) (D)}  &   0.769   & 0.724  & {0.758} & 0.537   & 0.622 & {0.550} \\
 &  \textbf{HASOC Best}  & -  & -  & \bftab{0.736} & -  & - & \bftab{0.575} \\
\bottomrule
\end{tabular}
\end{table}

Overall most of our models, show an improvement from the single models submitted at HASOC.The sub-par performance of the back-translated models across the sub-tasks suggest that data sparsity is not the central issue of this challenge. To take advantage of the augmented-data additional methods have to be used. The sub-task (D) does not significantly adds as an improvement to the models. It can be seen that it actually worsens the situation for sub-tasks B and sub-tasks C. This can be attributed to it changing the task to a much harder 7-class distribution task.The combined model approaches that we have mentioned above offer a resource efficient way for hate speech detection. The (ALL), (ALL) (MTL) and (MTL) models are able to generalize well for the different tasks and different languages. They present themselves as good candidates for a unified model for hate speech detection. 

\subsection{Error analysis}

\begin{figure}
    \centering
    \begin{subfigure}{\textwidth}
    \includegraphics[width=\linewidth]{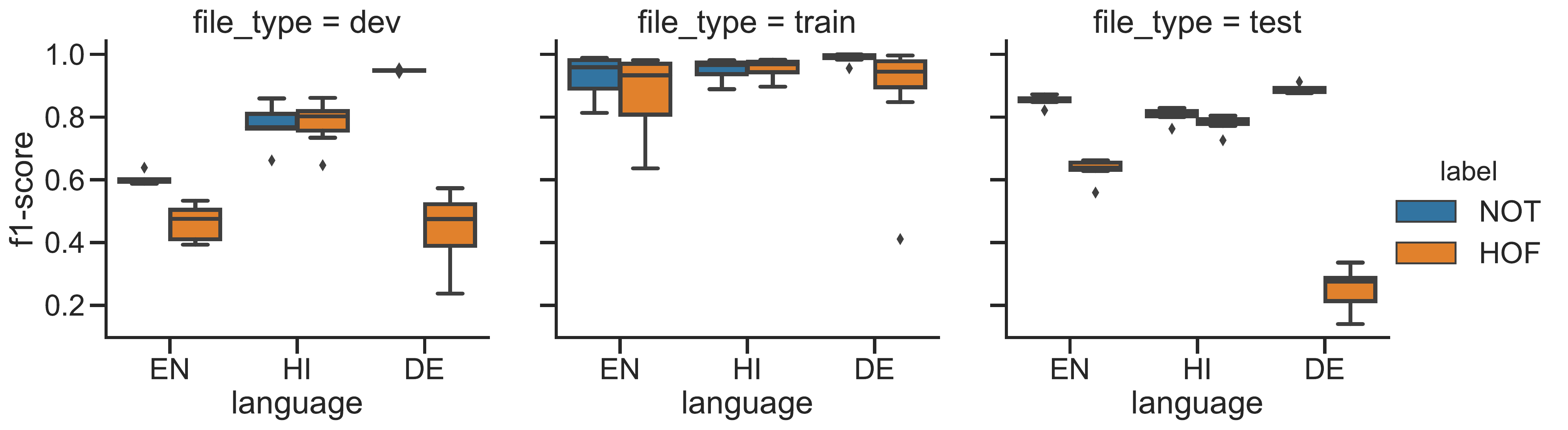}
    \caption{sub-task A}
    \label{fig:label_var:a}
    \end{subfigure}
    \begin{subfigure}{\textwidth}
    \includegraphics[width=\linewidth]{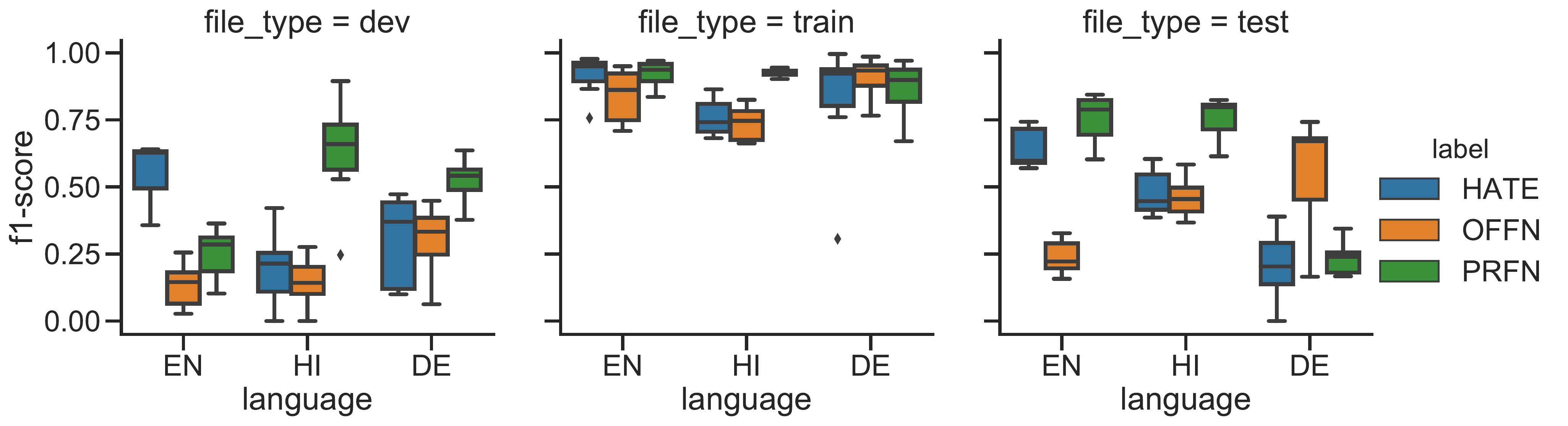}
    \caption{sub-task B}
    \label{fig:label_var:b}
    \end{subfigure}
    \begin{subfigure}{\textwidth}
    \includegraphics[width=\linewidth]{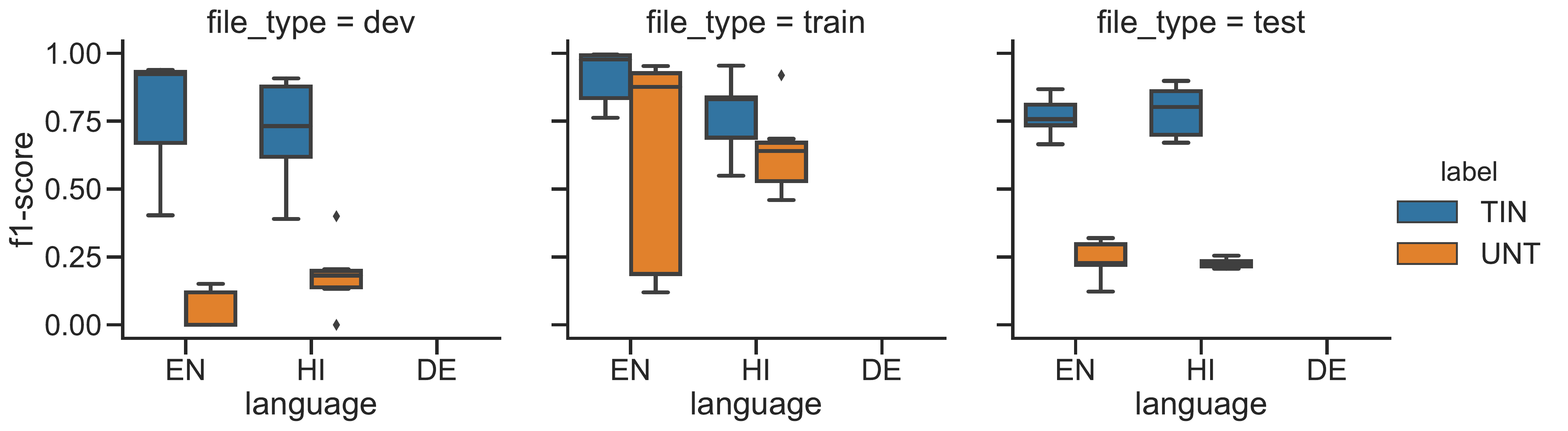}
    \caption{sub-task C}
    \label{fig:label_var:c}
    \end{subfigure}
    \caption{Variation in label F1-scores for all sub-tasks across all models}
    \label{fig:label_var}
\end{figure}

After identifying the best model and the variation in evaluation scores for each model, we investigate the overall performance of these models for each label belonging to each task. In Figure \ref{fig:label_var}, we can observe how the various labels have a high variance in their predictive performance.  

\subsubsection{Sub-Task A}
For English, the models show decent variation for both the labels on the training set. However, this variation is not as significant for the dev and test sets for the \textbf{(NOT)} label. There is an appreciable variation for \textbf{(HOF)} label on the dev set, but it is not transferred to the test set. The scores for the train sets is very high compared to the dev and test sets.
For Hindi, the predictions for both the labels show minimum variation in the F1-score across the three data-sets, with similar scores for each label.
For German, the F1-scores for the (NOT) class is quite high compared to that of \textbf{(HOF)} class. The models, have a very low F1-score for the \textbf{(HOF)} label on the test set with appreciable variation across the different models.

\subsubsection{Sub-Task B}
For English, the F1-scores for all the labels are quite high on the train set with decent variation for the \textbf{(OFFN)} label. All of the labels show appreciable variance on the dev and test sets. The \textbf{(OFFN)} has the lowest F1-score amongst the labels on both the dev and test sets with the other two labels having similar scores for the test set.
For Hindi, the train F1-scores are similar for all of the labels. The F1-scores are on the lower end for the \textbf{(HATE)} and \textbf{(OFFN)} labels on the dev set with appreciable variance across the models. This may be due to the fact that the Hindi dev set contains very few samples from these two labels.
For German, the variation among the F1-scores is high across all the three sets. The \textbf{(HATE)} label and the \textbf{(OFFN)} label have a large variation in their F1-scores across the models on the dev and test set respectively. The F1-score for the \textbf{(OFFN)} label is much higher than the other labels on the test set.

\subsubsection{Sub-Task C}
For English, the \textbf{(UNT)} label has exceptionally high variance across the models for the train set. This is due to the exceptionally low scores by the (BT) (ALL) (D) model. This label has extremely low F1-score on the dev set. Furthermore, there is also large variation in the \textbf{(TIN)} scores across the models in all the sets.

For Hindi, the \textbf{(TIN)} label has similar F1-scores with large variations across the models on all of the three sets. However, the \textbf{(UNT)} label has small variance across the models on the dev and test sets.

\subsubsection{Back Translation}
We also looked at the issue with back-translated results. In order to assess the back-translated data we looked at the new words added and removed from a sentence after back-translation. Aggregating these words over all sentences we find that the top words which are often removed and introduced are stop words, e.g. the, of, etc. In order to remove these stop words from our analysis and assess the salient words introduced and removed per label we remove the overall top 50 words from the introduced and removed list aggregated over each label. This highlights words which are often removed from offensive and hateful labels are indeed offensive words. A detailed list of words for English and German can be found in appendix \ref{appendix:back_translate} (we excluded results for Hindi because of Latex encoding issues).

\section{Discussion}

\subsection{Computational benefits}
From the results mentioned above, we can easily conclude that multi-task models present us with robust models for hate speech detection that can generalize well across different languages and different tasks for a given domain. Even the combined models, present us with models that can be deployed easily and give competitive performance on different languages with an efficient computation budget. Many of our models, perform better than the best scoring models of HASOC 2019 while maintaing a low inference cost.

\subsection{Additional evaluation}
Our current evaluation was limited to the HASOC dataset, additional evaluation needs to be done to assess out of domain and out of language capabilities of these techniques. Furthermore, the back-translation approach needs to be assessed even further using qualitative analysis of generated back-translations. 

\subsection{Architectures and Training improvements}
There are additional combination of architectures which we plan to try in future iterations of this work. Some of the combinations which we have not considered in this work are the (BT) (MTL) models and the (BT) (MTL) (ALL) models. We have seen that the (ALL) and (BT) models work well in unison and the (MTL) (ALL) models also give competitive performance with the (MTL) model. Therefore, a (BT) (MTL) (ALL) model is expected to bring out the best of both worlds. The (MTL) models we have used can still be tuned further, which may increase their results on the test sets. We trained (ALL) (MTL) model for 15 epochs instead of the usual 5, but it over-fitted the training set. Further experiments have to be conducted to identify the ideal training time for these models. 

\subsection{Real world usage}
Even though our models have performed really well on the HASOC dataset, yet the results are far from ideal. Given that HASOC dataset is quite small, our models may not generalize well outside of the domain of HASOC, however, our focus was on assessing the improvements we get using our multi-task and multi-lingual techniques on this datasets. We also conducted similar experiments in our work for the TRAC 2020 dataset \cite{Mishra2020TRAC}. In order to make these model more robust for general purpose hate-speech detection we need to train it on more-diverse and larger dataset. Furthermore, we also would like to highlight that our models need to be further evaluated for demographic bias as it has been found in \cite{Davidson2019} that hate speech and abusive language datasets exhibit racial bias towards African American English usage.

\section{Conclusion}

We would like to conclude this paper by highlighting the promise shown by multi-lingual and multi-task models on solving the hate and abusive speech detection in a computationally efficient way while maintaining comparable accuracy of single task models. We do highlight that our pre-trained models need to be further evaluated before being used on large scale, however the architecture and the training framework is something which can easily scale to large dataset without sacrificing performance as was shown in \cite{MishraThesisDSTDIE2020,Mishra2020ACMSigIR,Mishra2019MDMT}.  

\section*{Compliance with Ethical Standards}
Conflict of Interest: The authors declare that they have no conflict of interest.

\bibliographystyle{apalike}      
\bibliography{references}   

\appendix
\section*{Appendix}
\subsection{Label Distribution}
\label{appendix:label_dist}

\begin{figure}[!htb]
    \centering
    \begin{subfigure}[b]{\textwidth}
        \centering\
        \includegraphics[width=0.8\textwidth]{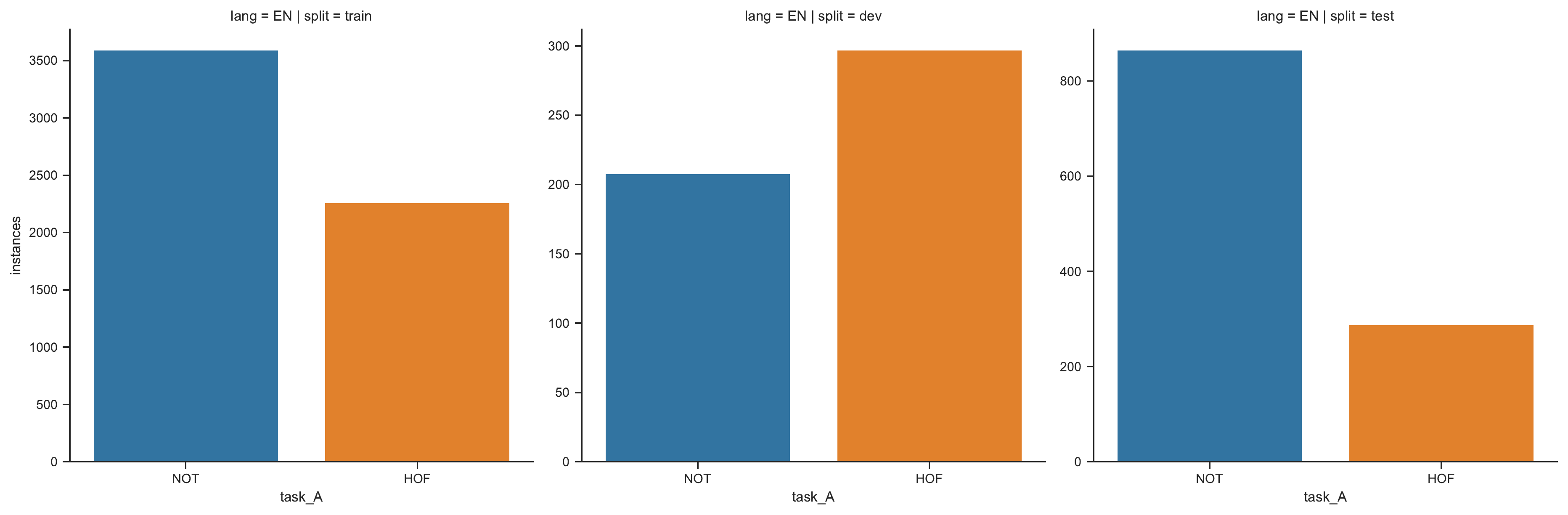}
    \end{subfigure}
    \begin{subfigure}[b]{\textwidth}
        \centering\
        \includegraphics[width=0.8\textwidth]{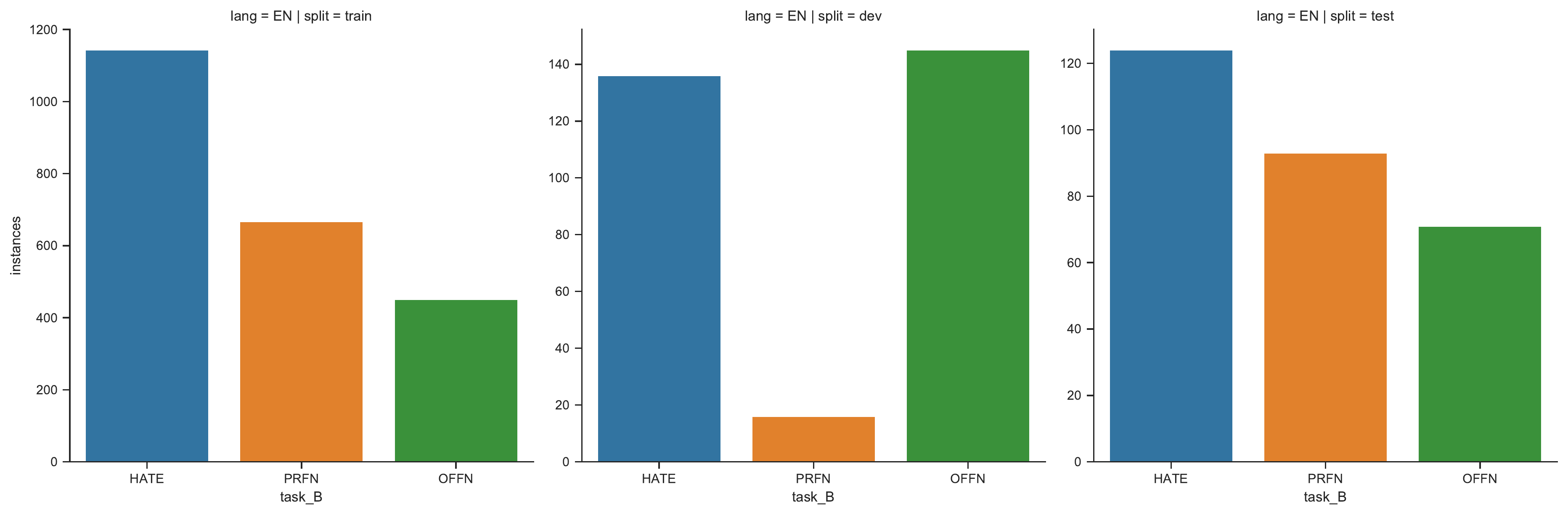}
    \end{subfigure}
    \begin{subfigure}[b]{\textwidth}
        \centering\
        \includegraphics[width=0.8\textwidth]{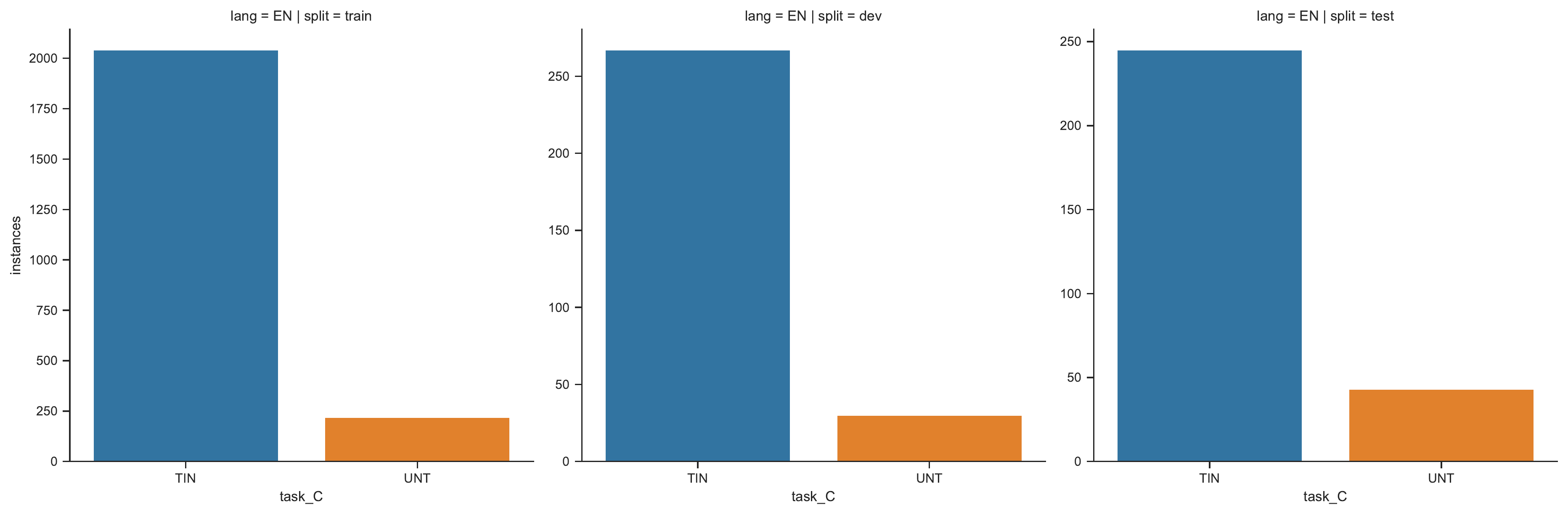}
    \end{subfigure}
    \caption{English Data class wise distribution }
    \label{fig:data_dist:a}
\end{figure}

\begin{figure}[!htb]
    \centering
    \begin{subfigure}[b]{\textwidth}
        \centering\
        \includegraphics[width=0.8\textwidth]{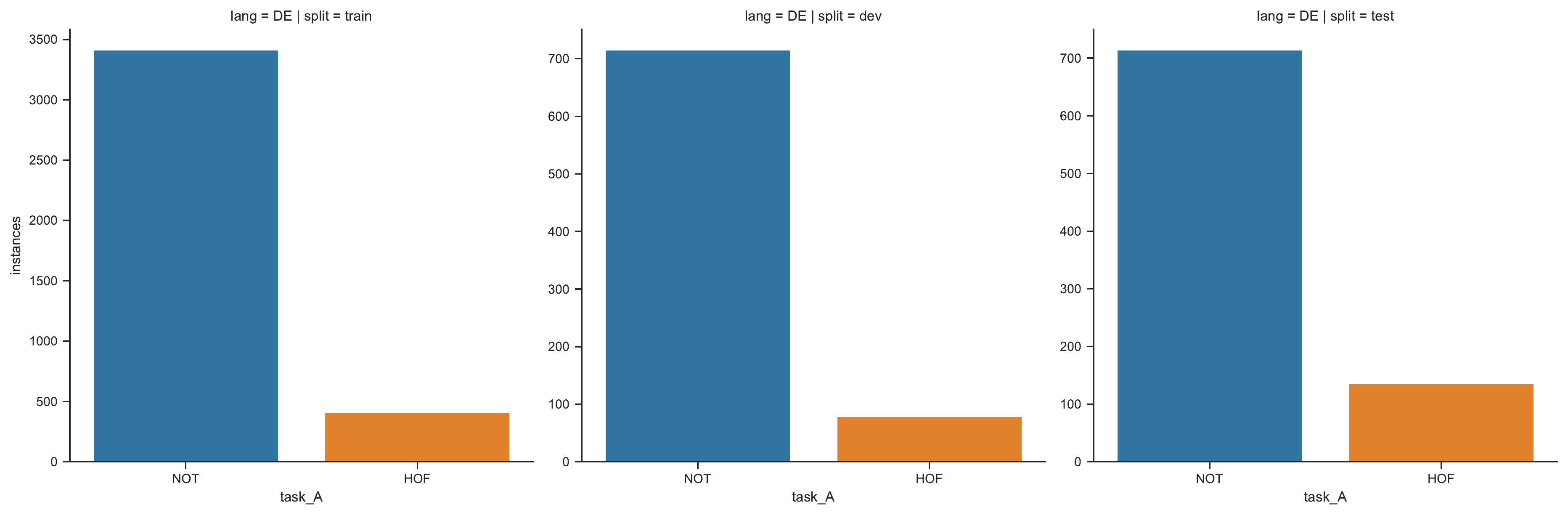}
    \end{subfigure}
    \begin{subfigure}[b]{\textwidth}
        \centering\
        \includegraphics[width=0.8\textwidth]{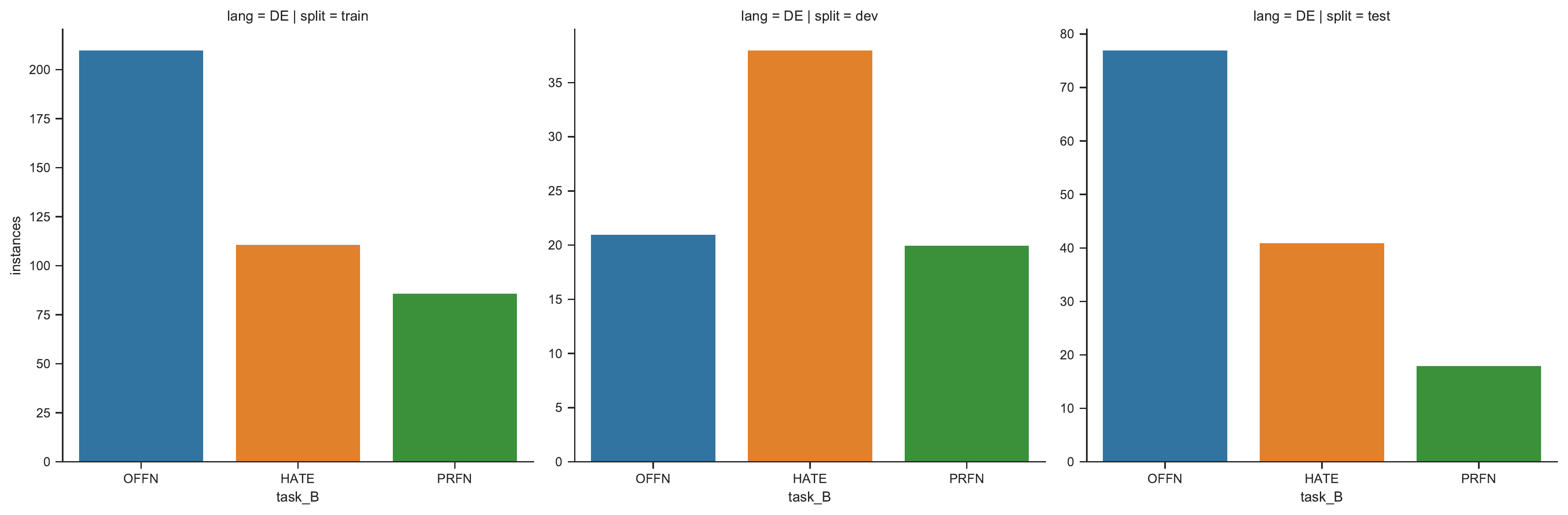}
    \end{subfigure}
    \caption{German Data class wise distribution }
    \label{fig:data_dist:b}
\end{figure}

\begin{figure}[!htb]
    \centering
    \begin{subfigure}[b]{\textwidth}
        \centering\
        \includegraphics[width=0.8\textwidth]{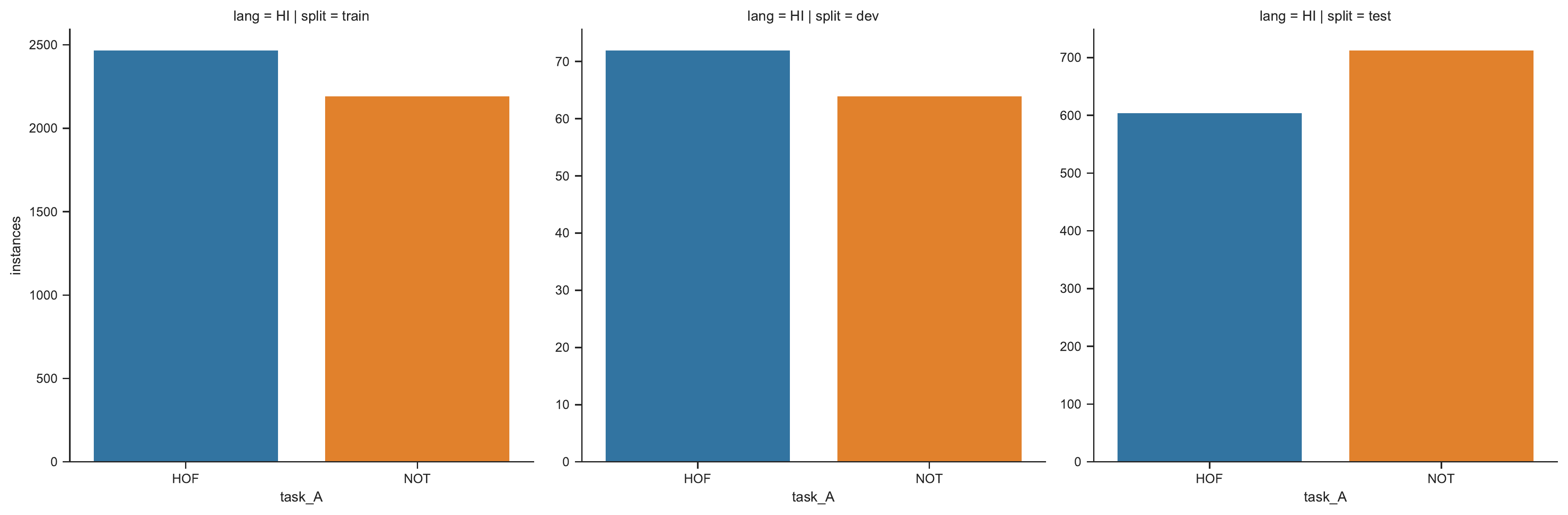}
    \end{subfigure}
    \begin{subfigure}[b]{\textwidth}
        \centering\
        \includegraphics[width=0.8\textwidth]{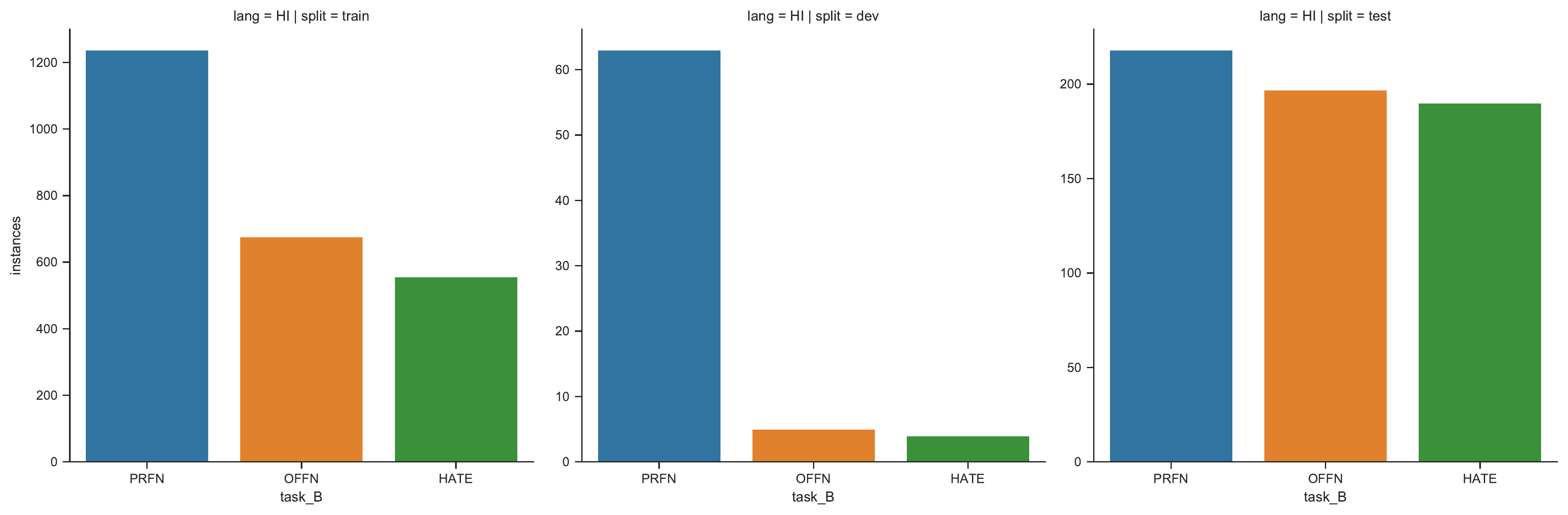}
    \end{subfigure}
    \begin{subfigure}[b]{\textwidth}
        \centering\
        \includegraphics[width=0.8\textwidth]{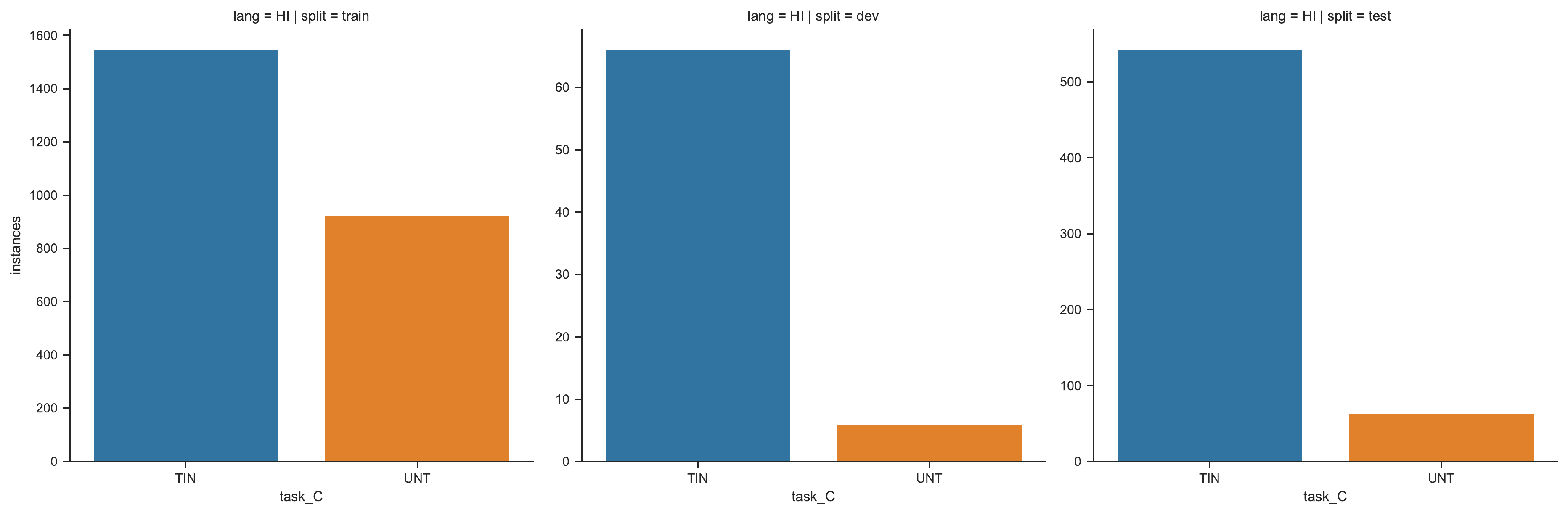}
    \end{subfigure}
    \caption{Hindi Data class wise distribution }
    \label{fig:data_dist:c}
\end{figure}

\clearpage

\subsection{Back translation top changed words}
\label{appendix:back_translate}

Here we list the top 5 words per label for each task obtained after removing the top 50 words which were either introduced or removed via back translation. We do not list the top words for Hindi because of the encoding issue in Latex.

\begin{lstlisting}[caption=Changed words in English Training Data,frame=single,breaklines,numbers=left]
task_1 introduced_words
HOF [('asset', 30), ("you're", 29), ('so', 28), ("it's", 26), ('there', 25)]
NOT [('worldcup2019', 47), ('at', 41), ("i'm", 40), ("it's", 38), ('us', 38)]
task_1 removed_words
HOF [('fuck', 52), ('he's', 43), ('what', 38), ('don't', 37), ('them', 36)]
NOT [('happy', 49), ('than', 47), ('being', 45), ('every', 45), ('been', 43)]
task_2 introduced_words
HATE [('there', 17), ('worldcup2019', 16), ('so', 14), ('because', 14), ('dhoni', 13)]
NONE [('worldcup2019', 47), ('at', 41), ("i'm", 40), ("it's", 38), ('us', 38)]
OFFN [('impeach45', 11), ('asset', 11), ('now', 8), ('lie', 6), ('trump2020', 6)]
PRFN [("it's", 16), ('fucking', 16), ('damn', 14), ('f', 14), ("you're", 12)]
task_2 removed_words
HATE [('which', 20), ('ground', 20), ('such', 18), ('its', 18), ("doesn't", 18)]
NONE [('happy', 49), ('than', 47), ('being', 45), ('every', 45), ('been', 43)]
OFFN [('he's', 12), ('them', 11), ("he's", 9), ('what', 9), ('been', 9)]
PRFN [('fuck', 47), ('fucking', 21), ('he's', 16), ('off', 16), ('don't', 15)]
task_3 introduced_words
NONE [('worldcup2019', 47), ('at', 41), ("i'm", 40), ("it's", 38), ('us', 38)]
TIN [('asset', 30), ("you're", 27), ('so', 25), ('which', 25), ('because', 25)]
UNT [('f', 8), ('nmy', 5), ('***', 5), ('there', 4), ('these', 4)]
task_3 removed_words
NONE [('happy', 49), ('than', 47), ('being', 45), ('every', 45), ('been', 43)]
TIN [('fuck', 48), ('he's', 39), ('what', 35), ('don't', 34), ("he's", 34)]
UNT [('them', 7), ('f***', 6), ('being', 5), ('such', 5), ('does', 5)]
\end{lstlisting}

\begin{lstlisting}[caption=Changed words in German Training Data,frame=single,breaklines,numbers=left]
task_1 introduced_words
HOF [('!!', 15), ('etwas', 15), ('diese', 12), ('sein', 11), ('werden', 11)]
NOT [('einen', 59), ('jetzt', 58), ('war', 57), ('menschen', 57), ('was', 56)]
task_1 removed_words
HOF [('du', 18), ('wohl', 14), ('haben', 12), ('eure', 11), ('mir', 11)]
NOT [('wieder', 56), ('uber', 55), ('vom', 52), ('haben', 51), ('einem', 49)]
task_2 introduced_words
HATE [('diese', 6), ('werden', 5), ('grun', 4), ('sollte', 4), ('konnen', 3)]
NONE [('einen', 59), ('jetzt', 58), ('war', 57), ('menschen', 57), ('was', 56)]
OFFN [('!!', 11), ('sein', 8), ('dumm', 8), ('etwas', 8), ('sein,', 7)]
PRFN [('ich', 6), ('scheibe', 5), ('etwas', 5), ('keine', 5), ('alle', 4)]
task_2 removed_words
HATE [('diesen', 5), ('dass', 5), ('kann', 4), ('wohl', 4), ('also', 4)]
NONE [('wieder', 56), ('uber', 55), ('vom', 52), ('haben', 51), ('einem', 49)]
OFFN [('du', 12), ('nur', 8), ('muss', 8), ('eure', 8), ('haben', 7)]
PRFN [('bin', 5), ('was', 5), ('wohl', 4), ('keine', 4), ('fressen', 4)]
\end{lstlisting}

\end{document}